\documentclass{article}

\IfFileExists{iclr2027_conference.sty}{%
  \usepackage{iclr2027_conference,times}%
}{%
  \usepackage{times}%
  \newcommand{\iclrfinalcopy}{}
}

\usepackage{amsmath,amssymb,amsfonts,bm}
\usepackage{booktabs}
\usepackage{graphicx}
\usepackage{multirow}
\usepackage{array}
\usepackage{xcolor}
\usepackage{enumitem}
\usepackage{natbib}
\usepackage{hyperref}
\hypersetup{hidelinks}
\usepackage{url}
\usepackage{microtype}
\usepackage{algorithm}
\usepackage{algorithmic}
\usepackage{tikz}
\usepackage[normalem]{ulem}   

\usetikzlibrary{arrows.meta,positioning,shapes.geometric,shapes.misc,fit,calc}

\newcommand{\method}{PRQuant}

\title{PRQuant: Permutation Residual Quantization for Low-Overhead Inference}

\author{%
  \normalfont
  Peiran Wang$^{1,2,*}$ \quad
  Anqi Wang$^{2,*}$ \quad
  Jiaying Zhao$^{1,*}$ \quad
  Huiwen Yang$^{1,\dagger}$ \quad
  Zhenyu Ming$^{1,\dagger}$ \\
  Yuantian Shao$^{3}$ \qquad
  Rongqian Wang$^{1}$ \qquad
  Yiwu Yao$^{1}$ \qquad
  Kun Tian$^{1}$ \qquad
  Xin Yao$^{1}$ \\
  Gong Zhang$^{1}$ \qquad
  Fan Yang$^{2}$ \qquad
  Zhongyi Huang$^{2}$
}

\iclrfinalcopy

\begin{document}

\maketitle

\begingroup
\renewcommand{\thefootnote}{}
\footnotetext{
\hspace{-1.8em}
$^{1}$Huawei Technologies Co., Ltd.
\qquad
$^{2}$Tsinghua University.
\qquad
$^{3}$Nanjing University of Science and Technology.
\\
$^{*}$Equal contribution.
\qquad
$^{\dagger}$Corresponding authors.
}
\endgroup

\begin{abstract}
Low-bit quantization of linear layers is often dominated by a small number of outlier channels. Existing smoothing, rotation, and residual-based methods can mitigate this issue, but may shift the quantization bottleneck to weights or introduce costly online operations. To address these limitations, we propose \method{} (Permutation Residual Quantization), a training-free framework that combines channel permutation with offline weight residual compensation. \method{} identifies the scaled-weight columns with the largest quantization errors and permutes them into contiguous tail blocks. This structure allows the corresponding weight residuals to be precomputed entirely offline, while replacing scattered activation gathering with simple contiguous access during inference, yielding a single regular MXFP4 GEMM for compensated computation.
Experiments show that \method{} substantially reduces down-projection reconstruction error, with scaling and residual compensation providing the main numerical gains while permutation enables a hardware-friendly contiguous layout.
Comprehensive experiment results on Qwen3-4B-Instruct-2507 and Qwen3-30B-A3B-Instruct-2507 illustrate that \method{} achieves up to averagelly $2.6\times$ and $1.8\times$ operator speedup over BF16 respectively, while preserving near plain MXFP4 end-to-end decoding efficiency. Across five downstream benchmarks, \method{} achieves the best average accuracy among the quantized methods, improving accuracy over MXFP4 by 1.24 and 0.55, respectively.
\end{abstract}

\section{Introduction}
\label{sec:introduction}

Large language models (LLMs), including mixture-of-experts (MoE)
architectures, have achieved strong performance in language understanding, code generation, and reasoning. However, their deployment is constrained by memory capacity, bandwidth, and the cost of large matrix multiplications~\citep{brown2020language,fedus2022switch, deepseek2025deepseekv3}. Low-bit quantization is therefore a central approach to reducing inference cost and improving throughput.

Recent hardware advances have increased interest in fine-grained block-scaled FP4 formats such as MXFP4 and NVFP4 ~\citep{rouhani2023microscaling,ocp2023mx}.
These formats share a scale within each small block, which confines an outlier’s influence to its local block. Within that block, however, a single large value can inflate the shared scale and coarsen the quantization of all other values. This unique block-wise duality implies that the properties of these formats need to be carefully considered in the design of quantization algorithms.

Existing post-training quantization (PTQ) strategies address outliers from different perspectives, but many established techniques were originally developed for integer quantization or settings without block-wise shared scaling.
Rotation-based methods use orthogonal transformations to redistribute activation outliers~\citep{ashkboos2024quarot}. Under MXFP4, global rotations may spread outlier energy across multiple scaling blocks and increase the
quantization error of otherwise regular
blocks~\citep{shao2025brq,sanjeet2026mixquant,li2026batquant}. BRQ and BATQuant align their transformations with MXFP4 blocks, but their activation-side transformations still run online during inference~\citep{shao2025brq,li2026batquant}.
SmoothQuant migrates quantization difficulty from activations to weights through a mathematically equivalent per-channel scaling transformation~\citep{xiao2023smoothquant}.
However, with weights also quantized to 4 bits, weight quantization can itself become the new bottleneck.
Residual compensation can partially recover the accuracy lost under aggressive 4-bit quantization, but typically introduces additional computation or memory-access overhead. For example, ARCQuant extends the activation matrix with quantized residual channels to preserve a unified low-precision GEMM path, but it depends on input-dependent residual channels~\citep{meng2026arcquant}. SVDQuant absorbs outliers using a low-rank branch, but this additional branch requires specialized kernel fusion to avoid extra data movement~\citep{li2024svdquant}.

\begin{figure}[t]
    \centering

    \hspace*{-0.08\linewidth}%
    \begin{minipage}[b]{0.49\linewidth}
        \centering
        \IfFileExists{prq1_a.png}{%
            \includegraphics[width=0.95\linewidth]{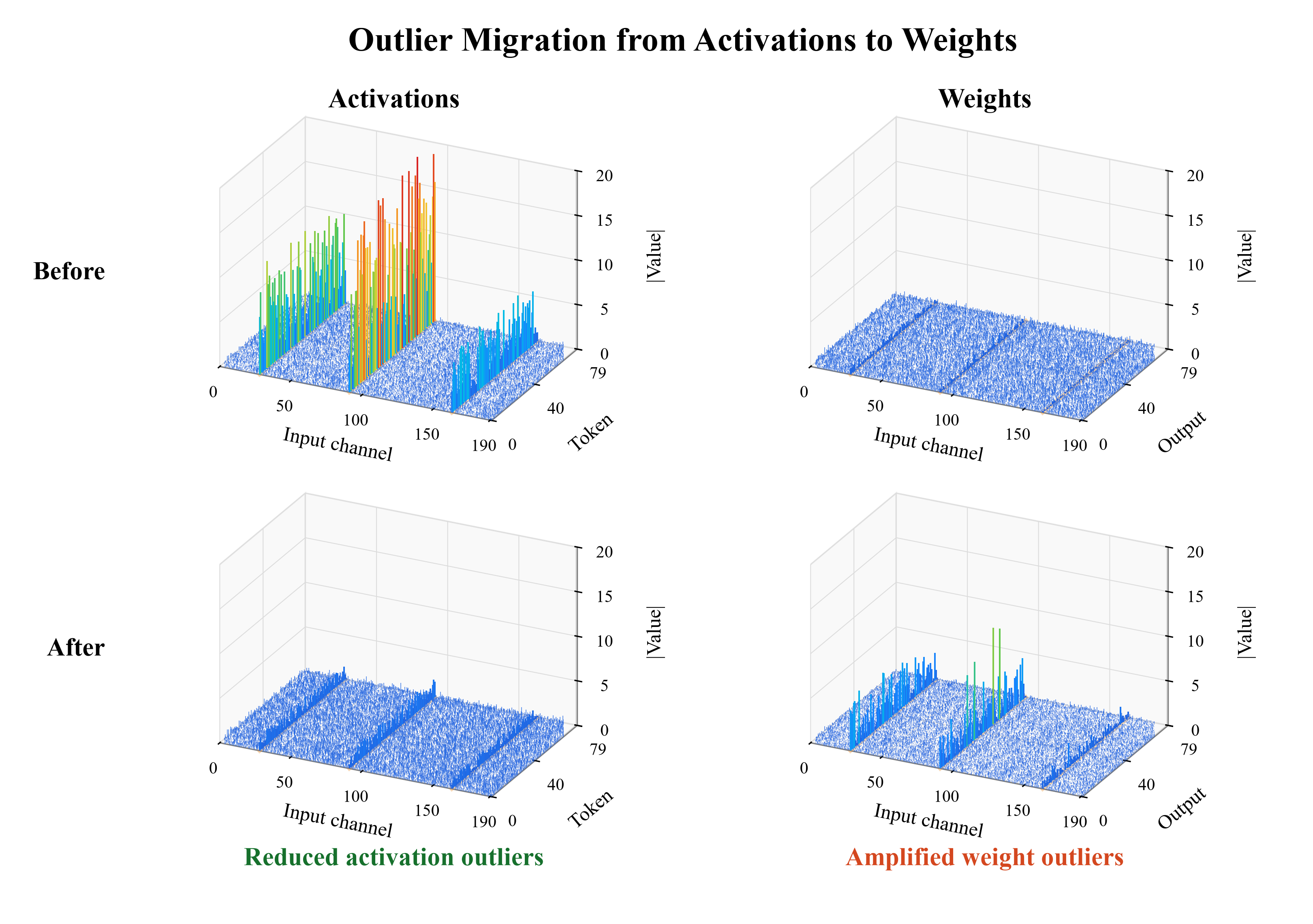}%
        }{%
            \fbox{\parbox[c][0.22\textheight][c]{0.94\linewidth}{%
                \centering Placeholder for \texttt{prq1_a.png}}}%
        }
    \end{minipage}%
    \hspace{0.02\linewidth}%
    \begin{minipage}[b]{0.5\linewidth}
        \centering
        \IfFileExists{prq1_b.png}{%
            \makebox[\linewidth][c]{%
                \includegraphics[width=1.2\linewidth]{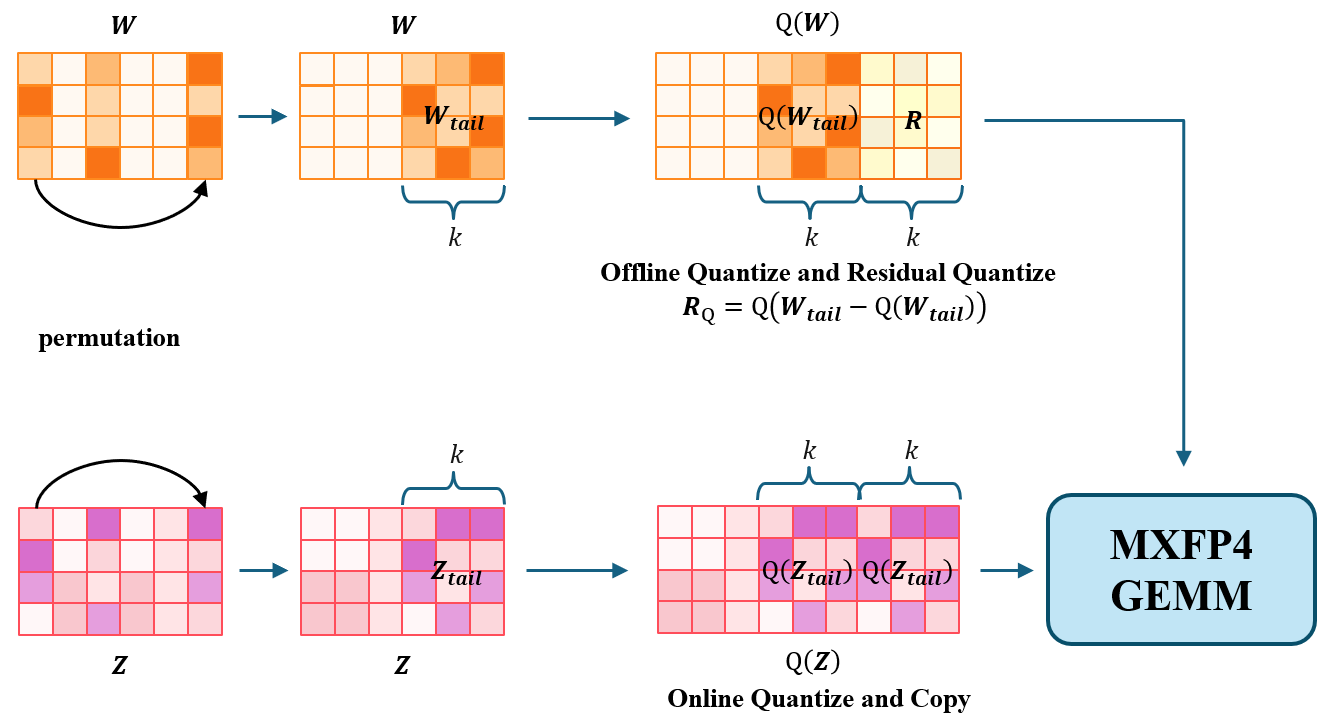}%
            }%
        }{%
            \fbox{\parbox[c][0.22\textheight][c]{0.94\linewidth}{%
                \centering Placeholder for \texttt{prq1_b.png}}}%
        }
    \end{minipage}%
    \par
    \vspace{3pt}

    \hspace*{-0.08\linewidth}%
    \begin{minipage}[t]{0.50\linewidth}
        \centering
        {\normalfont\normalsize (a)\par}
    \end{minipage}%
    \hspace{-0.02\linewidth}%
    \begin{minipage}[t]{0.50\linewidth}
        \centering
        {\normalfont\normalsize (b)\par}
    \end{minipage}%
    \par

    \caption{Schematic overview of PRQuant: (a) activation-to-weight quantization difficulty migration under scaling; (b) channel permutation, offline weight residual augmentation, and online activation augmentation.}
    \label{fig:workflow}
\end{figure}

These limitations motivate a low-overhead W4A4 method that preserves a regular low-precision GEMM path. We propose \textbf{PRQuant (Permutation Residual Quantization)}, whose overall pipeline is illustrated in Figure~\ref{fig:workflow}.
For each candidate $\alpha$, PRQuant first applies AWQ-style channel scaling and identifies the top-$k$ input channels that are most difficult to quantize in the scaled weights. These channels are permuted to the tail, where a static weight side residual sub-tensor is constructed from their MXFP4 quantization residuals. At inference time, PRQuant does not compute input-dependent activation residuals; it only reuses the corresponding tail activation channels, which can be obtained either by direct copying or through offline expansion of $W_{\mathrm{up}}$ and $W_{\mathrm{gate}}$. The main and residual paths can ultimately be merged into a single regular low-precision GEMM.

Our contributions are summarized as follows:
\begin{itemize}[leftmargin=*]
    \item We propose \method{} (Permutation Residual Quantization), a training-free PTQ framework that combines AWQ-style scaling, channel permutation, and offline weight residual compensation. The permutation moves the selected difficult channels into a contiguous tail region, so that the corresponding activations can be accessed efficiently during inference without scattered gathering.
    \item We design an efficient implementation of \method{} in which the selected weight residuals are precomputed offline and the main computation and residual compensation are executed together in a single augmented MXFP4 GEMM. This avoids online residual construction, irregular activation gathering, and a separate correction branch, while preserving a regular MXFP4 execution path.
    \item Extensive experiments on both dense and MoE Qwen3 models demonstrate that \method{} improves downstream accuracy over MXFP4 while modifying only the FFN down-projection layers. Our Ascend 950PR implementation further provides substantial down-projection speedups over BF16 and ARCQuant while retaining end-to-end decoding efficiency close to MXFP4, demonstrating both numerical effectiveness and hardware efficiency across distinct FFN architectures.
\end{itemize}

\section{Related Work}
\label{sec:related_work}

\paragraph{Scaling-based post-training quantization.}
SmoothQuant suppresses activation outliers by transferring quantization difficulty to weights through equivalent per-channel scaling~\citep{xiao2023smoothquant}. AWQ protects salient weights using activation-aware scaling~\citep{lin2024awq}, while OmniQuant optimizes learnable clipping and equivalent transformations~\citep{shao2024omniquant}. These methods are effective across low-bit quantization settings, but under W4A4, scaling may shift the bottleneck to equally constrained 4-bit weights.

\paragraph{Rotation and block-aware FP4 quantization.}
QuaRot and SpinQuant use randomized or learned rotations to reshape weight and activation distributions, while DartQuant further optimizes rotated activations~\citep{ashkboos2024quarot,liu2025spinquant,shao2025dartquant}. However, global rotations may redistribute outliers across MXFP4 blocks and increase block-wise quantization error~\citep{shao2025brq,li2026batquant}. Format-aware methods therefore operate at the microscaling granularity: MR-GPTQ combines GPTQ with block-wise rotations and scale optimization, while BRQ and BATQuant apply MXFP4-aligned transformations~\citep{egiazarian2025mrgptq,shao2025brq,li2026batquant}. These methods still require runtime activation-side transformations or fused-kernel support.

\paragraph{Channel reordering and outlier isolation.}
RPTQ reduces inter-channel range variation by clustering and reordering activation channels~\citep{yuan2023rptq}, while MixQuant uses permutation to redistribute activation mass across blocks~\citep{sanjeet2026mixquant}. These works show that channel layout affects both quantization behavior and execution efficiency, although reordering alone cannot fully recover the accuracy loss of 4-bit quantization.

\paragraph{Mixed-precision and residual compensation.}
LLM.int8() isolates outlier dimensions into a 16-bit path, while SpQR stores high-error weights in a sparse high-precision component~\citep{dettmers2022llmint8,dettmers2024spqr}. OSC similarly combines a 4-bit base GEMM with a 16-bit outlier branch~\citep{zhang2026osc}. ARCQuant avoids a high-precision branch by augmenting activations with quantized residual channels, but requires online residual construction~\citep{meng2026arcquant}; MosaicQuant combines dense and sparse 4-bit components with specialized execution support~\citep{hu2026mosaicquant}. These methods improve quantization accuracy through additional compensation capacity, but require extra runtime processing or execution support.

\section{Motivation}
\label{sec:motivation}

In practice, a small fraction of activation channels often contain large outliers (Figure~\ref{fig:workflow}(a)), which enlarge the quantization range and degrade 4-bit quantization accuracy. SmoothQuant alleviates this issue by transferring activation magnitudes to the weights through per-channel scaling. Although this transformation improves activation quantization accuracy, it may amplify a small number of scaled-weight input channels and create a new weight-side bottleneck under W4A4 quantization.

\paragraph{Limitations of high precision weight compensation.}
SVDQuant demonstrates that merely transferring quantization difficulty between activations and weights is insufficient at 4-bit precision. It applies a low-rank compensation approach to decompose the magnified weights into the sum of a 4-bit quantized weight matrix and a full-precision low-rank residual (LoRA)~\citep{li2024svdquant}. Although the low-rank component captures the dominant singular directions of the magnified weight and effectively preserves accuracy, it introduces significant overhead.
Executing the full-precision residual incurs extra data movement, whose efficient deployment relies on specialized mixed-precision kernel fusion. This makes the approach less compatible with a standard and homogeneous low-precision GEMM path.

\paragraph{Permutation enables structured weight compensation.}

After scaling, these dominant error channels are typically scattered across non-contiguous regions (Figure~\ref{fig:workflow}(b)), directly compensating for them would require expensive index gatherings. Therefore, we apply channel permutation to reorder these channels into contiguous tail blocks.  The permutation yields two benefits.  Numerically, it isolates the large-magnitude channels from regular ones, preventing outliers from inflating the shared quantization scales of otherwise well-behaved MXFP4 blocks. Structurally, it transforms the scattered indices into a contiguous data layout, thereby facilitating coalesced memory access and aligning well with hardware vectorized loading during inference.

\paragraph{Static weight residual compensation.}
Constructing residual compensation on the activation side is input-dependent and therefore requires online computation. In contrast, weight residuals are static and can be precomputed entirely offline. The remaining runtime requirement is only to provide the corresponding activation channels. AWQ-style scaling further shifts the dominant quantization difficulty to a small subset of weight channels, making weight-side residual compensation a natural way to target the remaining error. Meanwhile, permutation consolidates the corresponding channels into contiguous tail blocks, allowing their activations to be accessed without irregular gathering. Together, scaling and permutation make weight-side residual compensation both numerically targeted and lightweight at runtime, while retaining a regular MXFP4 GEMM path.

\section{Method}
\label{sec:method}

This section presents the \method{} pipeline for FFN down-projection layers (dense or MoE expert), as illustrated in Figure~\ref{fig:workflow}.

\paragraph{Offline scaling and channel reordering.}
Let $X$ denote the input to a down-projection layer. We use $Z$ to denote the intermediate activations of the gated FFN collected from a calibration set. Given a scaling strength $\alpha$, we construct the diagonal scaling matrix as
$D_{\alpha}=\operatorname{diag}(s_{\alpha})$, where
$s_{\alpha,j}=(\max_t |Z_{t,j}|)^\alpha$. We then identify the top-$k$ hardest-to-quantize channels in $W_{\mathrm{down}}D_{\alpha}$ and move them to contiguous tail blocks using a permutation matrix $P_{\alpha}$.

The scaling and permutation can be fused offline into the weights of adjacent linear layers. Specifically,
\begin{equation}
\begin{aligned}
\mathrm{MLP}(X)
&=
\left(
XW_{\mathrm{up}}^\top
\odot
\sigma(XW_{\mathrm{gate}}^\top)
\right)
W_{\mathrm{down}}^\top
\\
&=
\left(
X(P_{\alpha}^{\top}D_{\alpha}^{-1}W_{\mathrm{up}})^\top
\odot
\sigma\!\left(X(P_{\alpha}^{\top}W_{\mathrm{gate}})^\top\right)
\right)
(W_{\mathrm{down}}D_{\alpha}P_{\alpha})^\top
\\
&=
\widetilde Z_{\alpha}\widetilde W_{\mathrm{down},\alpha}^{\top}.
\end{aligned}
\label{eq:perm_sq_equiv}
\end{equation}
Here, the last $k$ columns of
$\widetilde Z_{\alpha}$ are denoted by
$\widetilde Z_{\alpha,\mathrm{tail}}$, and the corresponding columns of
$\widetilde W_{\mathrm{down},\alpha}$ are denoted by
$\widetilde W_{\mathrm{down},\alpha,\mathrm{tail}}$.

\paragraph{Offline residual augmentation.}
For the transformed down-projection weight in Eq.~\eqref{eq:perm_sq_equiv}, we compute the quantization residual of its tail channels as
\begin{equation}
R_{W,\alpha}
=
\widetilde W_{\mathrm{down},\alpha,\mathrm{tail}}
-
\mathcal Q\!\left(
\widetilde W_{\mathrm{down},\alpha,\mathrm{tail}}
\right),
\label{eq:weight_residual_alpha}
\end{equation}
where $\mathcal Q(\cdot)$ denotes the MXFP4 quantize-dequantize operator.

For notational simplicity, we omit the subscript $\alpha$ below. The residual is appended offline to the transformed down-projection weight, yielding
\begin{equation}
\begin{aligned}
\widehat Y
&=
\mathcal Q\!\left(
[\widetilde Z,\widetilde Z_{\mathrm{tail}}]
\right)
\mathcal Q\!\left(
[\widetilde W_{\mathrm{down}},R_{W}]
\right)^\top .
\end{aligned}
\label{eq:aug_weight_residual}
\end{equation}
Thus, the main computation and residual compensation are expressed as a single augmented MXFP4 GEMM whose reduction dimension increases from $C$ to $C+k$, where $C$ denotes the input dimension of the down-projection layer. For different candidate values of $\alpha$, we select the one that minimizes the normalized reconstruction error between $\widehat Y$ and $ZW_{\mathrm{down}}^\top$.

\paragraph{Activation tail construction.}
For the activation side of Eq.~\eqref{eq:aug_weight_residual}, we consider two implementations: online copy and offline copy.

\textit{Online copy.}
Since the permutation is fused into the adjacent projection weights, the selected channels already form a contiguous tail in $\widetilde Z$, i.e.,
$\widetilde Z_{\mathrm{tail}}=\widetilde Z[:,C-k:C]$. This contiguous layout avoids scattered gathering. The tail slice is then appended to the original activation, and $[\widetilde Z,\widetilde Z_{\mathrm{tail}}]$ is quantized as the input to the augmented GEMM.

\textit{Offline copy.}
Alternatively, activation duplication can be encoded in the producer weights during offline model preparation. Following Eq.~\eqref{eq:perm_sq_equiv}, the inverse channel scaling is absorbed into the up-projection weights, while the channel permutation is applied to both the up- and gate-projection weights. We then duplicate the last $k$ output channels of both transformed projections, so that the gated FFN directly produces $[\widetilde Z,\widetilde Z_{\mathrm{tail}}]$. This removes the runtime tail-copy operation at the cost of expanding the producer output dimension from $C$ to $C+k$.

\paragraph{Kernel design for efficient inference.}
\label{sec:kernel_design}
We implement \method{} on Ascend 950PR, following the online-copy formulation above. The quantized augmented weight matrix $
W_{\mathrm{aug}}^{q}
=
\mathcal Q\!\left(
[\widetilde W_{\mathrm{down}},R_{W}]
\right).$ are prepared offline. At runtime, the kernel quantizes the reordered activation $\widetilde Z$ to MXFP4 once, then appends a copy of its quantized tail together with the corresponding block scales. This reuses the quantized representation of $\widetilde Z_{\mathrm{tail}}$ without repeating quantization. Provided that the tail and concatenation boundaries are aligned with the MXFP4 quantization blocks and the copied blocks retain their original scales, this procedure is equivalent to quantizing $[\widetilde Z,\widetilde Z_{\mathrm{tail}}]$ in Eq.~\eqref{eq:aug_weight_residual}. A single MXFP4 GEMM then combines the augmented activation with $W_{\mathrm{aug}}^{q}$, extending the reduction dimension from $C$ to $C+k$ without online activation-residual construction or a separate correction GEMM.

\begin{algorithm}[t]
\caption{Calibration and Export Pipeline of \method{}}
\label{alg:prquant}
\small
\begin{algorithmic}[1]
\REQUIRE Down-projection weight $W_{\mathrm{down}}$, calibration activations $Z_{\rm cal}$, MXFP4 quantizer $\mathcal Q$, candidate set $\mathcal A$, tail width $k$
\STATE Compute the per-channel maxima of $Z_{\rm cal}$ for scaling.
\FOR{each $\alpha\in\mathcal A$}
    \STATE Construct $D_{\alpha}$ and the scaled weight $W_{\mathrm{down}}D_{\alpha}$.
    \STATE Select the top-$k$ channels with the largest column-wise MXFP4 quantization errors.
    \STATE Construct $P_{\alpha}$ to move the selected channels to a contiguous tail block.
    \STATE Obtain $\widetilde Z_{\alpha}$ and $\widetilde W_{\mathrm{down},\alpha}$ using Equation~\eqref{eq:perm_sq_equiv}.
    \STATE Compute
    $R_{W,\alpha}
    =
    \widetilde W_{\mathrm{down},\alpha,\mathrm{tail}}
    -
    \mathcal Q(\widetilde W_{\mathrm{down},\alpha,\mathrm{tail}})$.
    \STATE Construct
    $[\widetilde Z_{\alpha},\widetilde Z_{\alpha,\mathrm{tail}}]$
    and
    $[\widetilde W_{\mathrm{down},\alpha},R_{W,\alpha}]$,
    and evaluate Equation~\eqref{eq:aug_weight_residual}.
\ENDFOR
\STATE Select $\alpha^\star$ with the minimum reconstruction error and fuse the corresponding $(P_{\alpha^{*}},D_{\alpha^{*}})$ into the up-, gate-, and down-projection weights.
\STATE Export the augmented down-projection weight
$[\widetilde W_{\mathrm{down}},R_W]$ in MXFP4.
\STATE For online copy, quantize $\widetilde Z$ once and reuse its quantized contiguous tail; alternatively, duplicate the corresponding producer channels offline.
\STATE Execute one MXFP4 GEMM with the augmented activation and weight.
\end{algorithmic}
\end{algorithm}

\section{Experiments}
\label{sec:experiments}

\subsection{Experimental Setup}
\label{sec:exp_setup}

\paragraph{Model and quantization configuration.}
We use Qwen3-4B-Instruct-2507 and Qwen3-30B-A3B-Instruct-2507 \citep{yang2025qwen3} as our primary evaluation models, denoted as Qwen3-4B and Qwen3-30B hereafter. Unless otherwise specified, \method{} is applied only to the FFN down-projection layers: the dense FFN down-projections for Qwen3-4B and the expert down-projections for Qwen3-30B. All compared methods use the same model checkpoint and evaluation pipeline. We evaluate a mixed MXFP quantization configuration, where the attention modules are quantized in MXFP8 and the FFN/MoE modules are quantized in MXFP4.

\paragraph{Baseline methods.}
We benchmark \method{} against full-precision BF16, default MXFP4, and five representative low-bit quantization methods. Specifically, BF16 provides the full-precision reference, whereas default MXFP4 directly quantizes both activations and weights without additional quantization-aware transformations or compensation. The low-bit baselines are: ARCQuant, which compensates for activation quantization error; BATQuant, which learns block-wise affine transformations and clipping~\citep{li2026batquant}; FlatQuant, which learns affine transformations to reshape weight and activation distributions~\citep{flatquant}; MR-GPTQ, which combines Hessian-based weight quantization with MXFP-aligned rotations and scale optimization~\citep{egiazarian2025mrgptq}; and QuaRot, for which we apply random Hadamard rotations to the down-projection layers~\citep{ashkboos2024quarot}. All quantized methods follow the same mixed-precision setting, with MXFP8 for the attention modules and MXFP4 for the FFN/MoE modules.

\paragraph{Calibration and benchmarking datasets.}
All quantization methods use the same calibration set, \texttt{mix\_calib.jsonl} \footnote{\url{http://gitcode.com/Ascend/msmodelslim/blob/master/lab_calib/mix_calib.jsonl}}. We evaluate end-to-end accuracy across five standard benchmarks: ARC-Challenge (ARC-C), ARC-Easy (ARC-E),
HellaSwag, PIQA, and WinoGrande
\citep{clark2018think,zellers2019hellaswag,bisk2020piqa,sakaguchi2021winogrande}.

\paragraph{Implementation and environmental setup.}
For the end-to-end accuracy comparison, all methods are implemented and evaluated under identical environmental conditions on the Ascend 950PR.
Specifically, all model checkpoints (including full-precision and quantized results) are deployed using the \texttt{vllm-ascend (0.20.2 RC1)} serving framework, and end-to-end evaluation is conducted uniformly via OpenCompass~\citep{opencompass2023}.

\subsection{End-to-End Accuracy}
\label{sec:e2e_accuracy}

\begin{table}[t]
\centering

\caption{
End-to-end accuracy ($\uparrow$) on Qwen3-30B and Qwen3-4B using zero-shot evaluation.
For quantized configurations, the attention modules use MXFP8 and the FFN/MoE modules use MXFP4.
Bold values denote the best result among the quantized configurations for each model.
The average is the unweighted mean over the five benchmarks.
}
\label{tab:e2e_accuracy}

\vspace{6pt} 

\begin{tabular}{l|c c c c c|c}
\toprule
Method & ARC-C & ARC-E & HellaSwag & PIQA & WinoGrande & Average \\
\midrule
\multicolumn{7}{c}{\textbf{Qwen3-30B}} \\
\cmidrule(lr){1-7}
BF16      & 95.93 & 98.77 & 90.19 & 92.27 & 75.92 & 90.62 \\
MXFP4     & 94.57 & 98.58 & 88.74 & 90.81 & 74.03 & 89.35 \\
ARCQuant  & \textbf{95.25} & \textbf{98.77} & 87.23 & 91.57 & 73.01 & 89.17 \\
BATQuant  & 94.24 & 98.59 & 88.60 & 91.62 & 74.59 & 89.53 \\
FlatQuant & 94.92 & \textbf{98.77} & 88.73 & 91.51 & 75.14 & 89.81 \\
MR-GPTQ   & 94.23 & 97.88 & \textbf{89.21} & 91.46 & 75.30 & 89.62 \\
QuaRot    & 94.92 & 97.88 & 87.70 & 91.51 & 72.38 & 88.88 \\
\method{} & 94.58 & \textbf{98.77} & 88.77 & \textbf{91.95} & \textbf{75.45} & \textbf{89.90} \\
\midrule
\multicolumn{7}{c}{\textbf{Qwen3-4B}} \\
\cmidrule(lr){1-7}
BF16      & 93.22 & 97.71 & 80.56 & 85.26 & 63.61 & 84.07 \\
MXFP4     & 90.85 & 96.65 & 73.83 & 83.08 & \textbf{60.38} & 80.96 \\
ARCQuant  & 92.20 & 96.65 & 75.28 & 83.95 & 58.56 & 81.33 \\
BATQuant  & 92.20 & 96.47 & 77.30 & \textbf{84.44} & 59.51 & 81.98 \\
FlatQuant & 90.17 & 95.23 & \textbf{77.96} & 83.79 & 59.98 & 81.43 \\
MR-GPTQ   & 89.83 & \textbf{97.18} & 76.41 & 83.30 & 59.27 & 81.20 \\
QuaRot    & 90.85 & 97.00 & 75.32 & 83.35 & 59.59 & 81.22 \\
\method{} & \textbf{92.54} & \textbf{97.18} & 77.51 & 83.84 & 59.91 & \textbf{82.20} \\
\bottomrule
\end{tabular}

\end{table}

We report the end-to-end accuracy in Table~\ref{tab:e2e_accuracy}.
For Qwen3-30B, default MXFP4 reduces the average accuracy from 90.62 to 89.35, corresponding to a 1.27 degradation from BF16. By modifying only the expert down-projection computation, \method{} improves the average accuracy to 89.90, yielding a 0.55 gain over MXFP4 and recovering approximately 43.3\% of the accuracy gap between MXFP4 and BF16. 

The improvement over MXFP4 is observed across all five benchmarks. In particular, \method{} improves WinoGrande and PIQA by 1.42 and 1.14, respectively, while providing smaller gains of 0.19, 0.03, and 0.01 on ARC-E, HellaSwag, and ARC-C. Compared with the BF16 reference, \method{} fully preserves the ARC-E accuracy and limits the PIQA degradation to only 0.32. These empirical results demonstrate that mitigating the reconstruction error in sensitive expert down-projections improves end-to-end accuracy.

Although their average accuracies are slightly lower than that of \method{}, BATQuant, FlatQuant, and MR-GPTQ achieve comparable results on this model. We therefore compare their quantization cost in Table~\ref{tab:time_mem_30b}. The training-based BATQuant and FlatQuant and the Hessian-based MR-GPTQ require 7.4--9.1 hours, whereas the training-free \method{} completes quantization in 0.4 hours and uses less memory. These results make \method{} the most economical choice among the compared methods for quantizing Qwen3-30B, combining comparable or better accuracy with the lowest time and memory costs. 

For the dense Qwen3-4B, the quantization gap is substantially larger: default MXFP4 reduces the average accuracy from 84.07 to 80.96, corresponding to a 3.11 degradation from BF16. Applying \method{} to the dense FFN down-projections improves the average accuracy to 82.20, yielding a 1.24 improvement over MXFP4 and recovering approximately 39.9\% of the average accuracy gap between MXFP4 and BF16. \method{} again achieves the highest average accuracy among the evaluated quantized methods.

Compared with MXFP4, \method{} improves four of the five benchmarks, with gains of 1.69 on ARC-C, 0.53 on ARC-E, 3.68 on HellaSwag, and 0.76 on PIQA. The largest improvement is observed on HellaSwag, suggesting that the dense model is particularly sensitive to quantization errors in the FFN down-projections. Although individual baselines remain competitive on specific tasks, \method{} provides the strongest overall accuracy across the five benchmarks. These results show that reducing reconstruction error in the targeted down-projections translates into end-to-end accuracy gains for both MoE expert FFNs and dense FFNs, while requiring no modification to the attention computation.

\subsection{Ablation Studies}
\label{sec:ablation_studies}

\begin{figure}[t]
    \centering
    \small

    \begin{minipage}[c]{\linewidth}
        \centering

        \vspace{2pt}

        \includegraphics[width=0.65\linewidth]{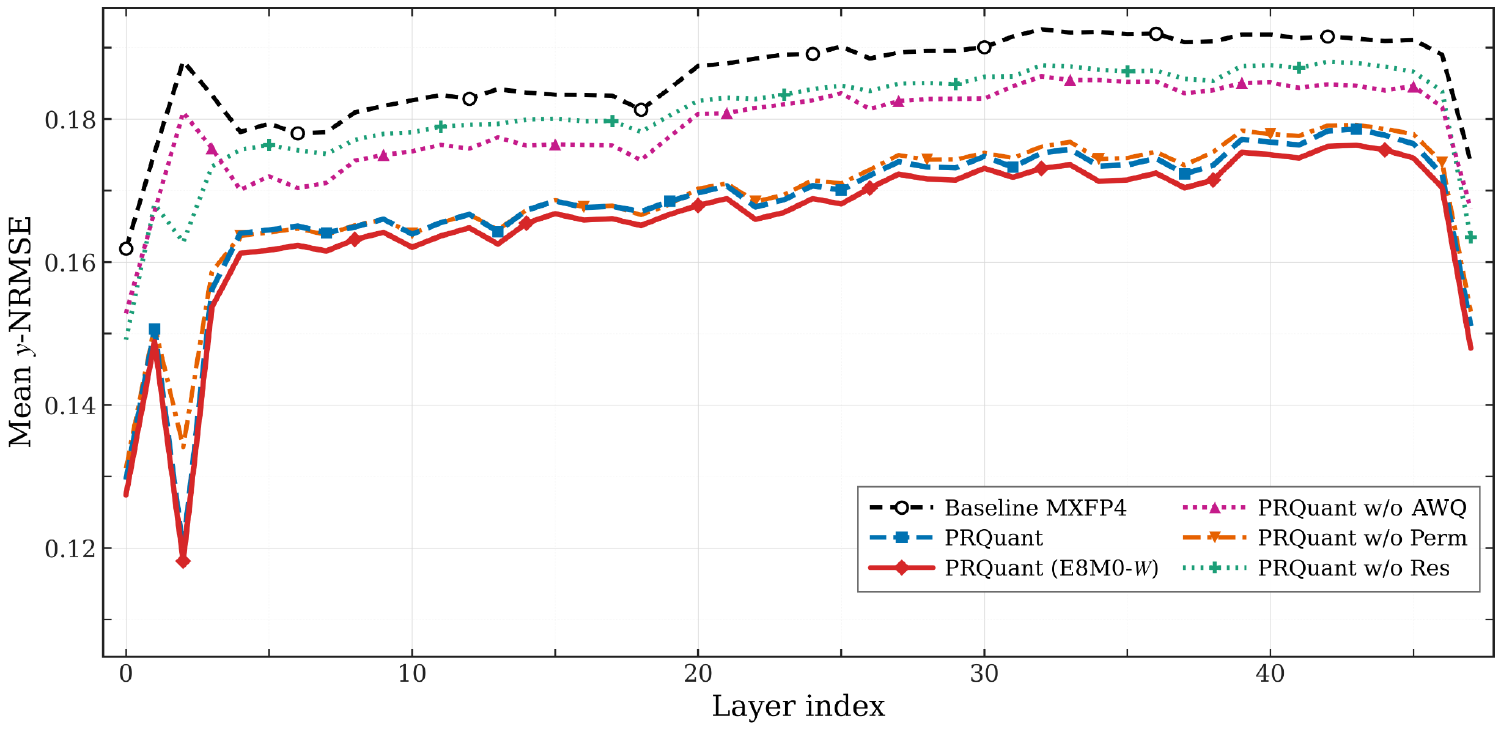}
    \end{minipage}

    \caption{
    Ablation results on the down-projection layers of routed experts of Qwen3-30B.
    Layer-wise mean $y$-NRMSE across experts.
    }
    \label{fig:ablation}
\end{figure}

We conduct layer-wise ablations on the expert down-projection layers of
Qwen3-30B. We report the mean normalized reconstruction error (denoted as $y$-NRMSE) across experts in each layer. We compare the complete \method{} with naive MXFP4 quantization and three ablated variants that remove AWQ scaling (w/o AWQ), residual compensation (w/o Res), and channel permutation (w/o Perm), respectively. We additionally evaluate \method{} with weight-side E8M0 block-scale search, denoted as \method{} (E8M0-W). 
Figure~\ref{fig:ablation} presents the averaged $y$-NRMSE across all evaluated layers and the layer-wise results.

\method{} reduces the average $y$-NRMSE from 0.1861(naive MXFP4) to 0.1678.
Removing AWQ scaling or residual compensation causes the largest
degradation, showing that they provide the main numerical gains.
Removing permutation results in only a modest error increase, while its
primary benefit is structural: it reorganizes scattered hard channels
into a contiguous tail block, enabling residual compensation through a
regular augmented GEMM without dynamic gathering.
Finally, introducing a local block-scale search \method{} (E8M0-W) further suppresses the error to 0.1658, demonstrating that optimizing weight scaling factors can achieve even higher reconstruction accuracy under MXFP4 quantization.

\subsection{Efficiency and Speed}
\label{sec:efficiency}

We evaluate the runtime efficiency of \method{} on Ascend 950PR at both the operator and end-to-end levels. We compare BF16, plain MXFP4, ARCQuant, and \method{} under the same inference configuration. For end-to-end measurements, we use identical output lengths, KV-cache precision, and KV-cache capacity across all methods.

\begin{figure}[t]
    \centering
    \begin{minipage}{0.48\linewidth}
        \centering
        \includegraphics[width=\linewidth]{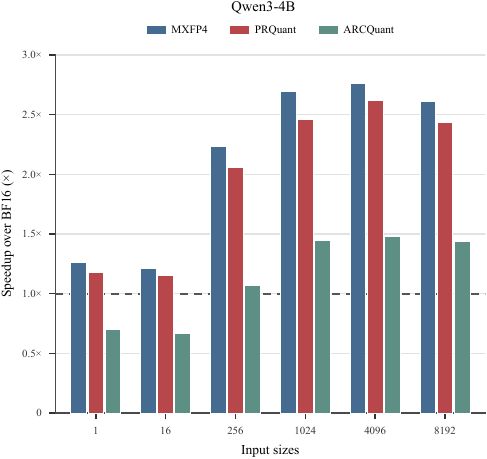}
    \end{minipage}
    \hfill
    \begin{minipage}{0.48\linewidth}
        \centering
        \includegraphics[width=\linewidth]{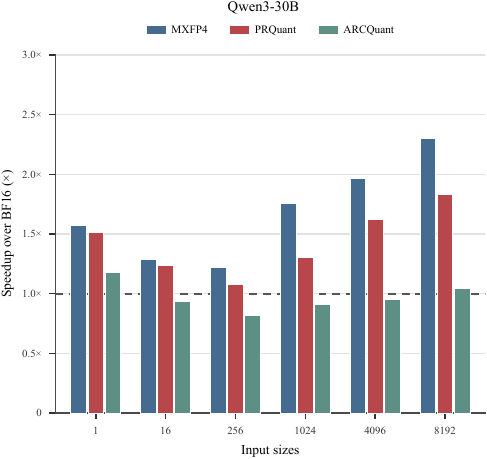}
    \end{minipage}
    \caption{Operator-level Speedup ($\uparrow$) over BF16 on Ascend 950PR for the down-projection kernels of Qwen3-4B and Qwen3-30B across representative input sizes $M$.
    The dashed horizontal line denotes the BF16 baseline ($1\times$).}
    \label{fig:operator_speedup}
\end{figure}

\paragraph{Operator efficiency.}
Figure~\ref{fig:operator_speedup} compares the speedups of the down-projection kernels relative to BF16 across input sizes $M$. On Qwen3-4B and Qwen3-30B, \method{} achieves up to approximately $2.6\times$ and $1.8\times$ speedup over BF16, respectively, while being about $1.8\times$ and $1.6\times$ faster than ARCQuant across the representative workloads. The additional advantage over ARCQuant mainly comes from the lightweight online activation handling. Compared with plain MXFP4, \method{} introduces additional computation from the expanded GEMM reduction dimension and the tail copy operation, and therefore naturally incurs some latency overhead. Nevertheless, \method{} incurs only about $5$-$10\%$ additional latency on Qwen3-4B, while the gap is larger on Qwen3-30B. The larger gap on Qwen3-30B is consistent with its larger relative reduction dimension expansion: $k/C=16.7\%$ ($C=768$, $k=128$), compared with $10.5\%$ ($C=9728$, $k=1024$) on Qwen3-4B. Moreover, the smaller expert projections in Qwen3-30B take less time to compute, so tail copying and associated activation-processing overhead account for a larger fraction of the total kernel latency.


\paragraph{End-to-end efficiency.}
Tables~\ref{tab:ttft_e2e} and~\ref{tab:tpot_e2e} report the end-to-end TTFT and TPOT results. \method{} consistently reduces both TTFT and TPOT relative to BF16 across all tested configurations. During decoding, its latency remains within approximately $6\%$ of plain MXFP4 on Qwen3-4B and $4\%$ on Qwen3-30B. Compared with ARCQuant, \method{} reduces TTFT by approximately $4.5$-$38.8\%$ and TPOT by $1.8$-$6.8\%$ on Qwen3-4B.
On Qwen3-30B, \method{} achieves up to $14.9\%$ lower TTFT than
ARCQuant while preserving decode efficiency, with the TPOT difference
remaining within $1.6\%$. Together with the accuracy results, these results show that \method{} preserves most of the decoding efficiency of plain MXFP4 while providing improved quantization accuracy.

\begin{table}
    \centering
    \caption[Quantization cost of Qwen3-30B]{Quantization cost ($\downarrow$) of Qwen3-30B. Both BATQuant and FlatQuant are implemented using the \texttt{ModelSlim}\protect\footnotemark[2] framework.}
    \label{tab:time_mem_30b}
    \vspace{6pt}
    \begin{tabular}{c|cc}
        \toprule
        Method & Quantization Latency (h) & Memory (GiB) \\
        \midrule
        BATQuant  & 8.5  & 36.1 \\
        FlatQuant & 7.4 & 22.9 \\
        MR-GPTQ   & 9.1 & 59.2 \\
        \method{}   & 0.4 & 4.4  \\
        \bottomrule
    \end{tabular}
    
\end{table}
\footnotetext[2]{\url{https://gitcode.com/Ascend/msmodelslim}}


\begin{table*}[t]
\centering
\caption{
Prefill latency TTFT (ms, $\downarrow$) of Qwen3-4B and Qwen3-30B on Ascend 950PR
under different batch sizes and input sequence lengths (ISL).
}
\label{tab:ttft_e2e}
\vspace{6pt}
\small
\setlength{\tabcolsep}{4.5pt}
\renewcommand{\arraystretch}{1.08}

\begin{tabular}{cc|cccc|cccc}
\toprule
\multirow{2}{*}{Batch}
& \multirow{2}{*}{ISL}
& \multicolumn{4}{c|}{Qwen3-4B}
& \multicolumn{4}{c}{Qwen3-30B} \\
\cmidrule(lr){3-6}
\cmidrule(lr){7-10}
&
& BF16
& MXFP4
& \method{}
& ARCQuant
& BF16
& MXFP4
& \method{}
& ARCQuant \\
\midrule

\multirow{3}{*}{1}
& 1024
& 41.2 & 30.8 & 31.8 & 52.0
& 65.5 & 52.8 & 55.7 & 62.9 \\
& 2048
& 72.2 & 47.0 & 46.1 & 54.1
& 92.5 & 65.8 & 86.8 & 85.1 \\
& 4096
& 147.8 & 87.4 & 88.0 & 93.2
& 186.2 & 129.8 & 161.7 & 166.5 \\
\midrule

\multirow{3}{*}{8}
& 1024
& 224.9 & 132.4 & 132.8 & 171.2
& 250.4 & 174.6 & 234.6 & 248.7 \\
& 2048
& 346.4 & 189.6 & 192.5 & 222.5
& 458.8 & 296.5 & 411.3 & 447.9 \\
& 4096
& 640.6 & 390.7 & 395.1 & 424.1
& 917.8 & 610.2 & 749.5 & 847.3 \\
\midrule

\multirow{3}{*}{16}
& 1024
& 346.2 & 198.9 & 202.0 & 237.3
& 430.0 & 277.6 & 383.2 & 401.7 \\
& 2048
& 628.2 & 352.0 & 352.6 & 398.2
& 826.4 & 529.8 & 716.8 & 757.8 \\
& 4096
& 1228.0 & 776.8 & 782.8 & 833.2
& 1746.8 & 1168.8 & 1334.1 & 1561.5 \\
\midrule

\multirow{3}{*}{32}
& 1024
& 595.4 & 338.8 & 344.0 & 385.5
& 767.5 & 489.6 & 658.7 & 696.3 \\
& 2048
& 1179.1 & 667.3 & 684.7 & 757.3
& 1560.6 & 1004.3 & 1265.8 & 1389.6 \\
& 4096
& 2427.7 & 1599.9 & 1610.9 & 1686.9
& 3411.0 & 2314.5 & 2545.4 & 2992.5 \\
\bottomrule
\end{tabular}


\end{table*}

\begin{table}[t]
\centering
\caption{
Decode latency TPOT (ms/token, $\downarrow$) and speedup ($\uparrow$) over BF16 of Qwen3-4B and Qwen3-30B on Ascend 950PR
with $\mathrm{ISL}=2048$ and $\mathrm{OSL}=128$.
}
\label{tab:tpot_e2e}
\vspace{6pt}
\setlength{\tabcolsep}{4.5pt}
\renewcommand{\arraystretch}{1.08}

\begin{tabular}{cc|cccc|ccc}
\toprule
\multirow{2}{*}{Model}
& \multirow{2}{*}{Batch}
& \multicolumn{4}{c|}{TPOT (ms/token)}
& \multicolumn{3}{c}{Speedup} \\
\cmidrule(lr){3-6}
\cmidrule(lr){7-9}
&
& BF16
& MXFP4
& \method{}
& ARCQuant
& MXFP4
& \method{}
& ARCQuant \\
\midrule

\multirow{4}{*}{Qwen3-4B}
& 1
& 7.4
& 4.9
& 5.2
& 5.3
& 1.51$\times$
& 1.42$\times$
& 1.40$\times$ \\

& 8
& 11.0
& 7.7
& 7.7
& 8.2
& 1.43$\times$
& 1.43$\times$
& 1.34$\times$ \\

& 16
& 15.3
& 11.0
& 11.0
& 11.8
& 1.39$\times$
& 1.39$\times$
& 1.30$\times$ \\

& 32
& 23.9
& 17.9
& 17.9
& 19.0
& 1.34$\times$
& 1.34$\times$
& 1.26$\times$ \\

\midrule

\multirow{4}{*}{Qwen3-30B}
& 1
& 8.2
& 6.6
& 6.8
& 6.9
& 1.24$\times$
& 1.21$\times$
& 1.19$\times$ \\

& 8
& 24.7
& 13.4
& 13.7
& 13.7
& 1.84$\times$
& 1.80$\times$
& 1.80$\times$ \\

& 16
& 25.1
& 15.2
& 15.6
& 15.4
& 1.65$\times$
& 1.61$\times$
& 1.63$\times$ \\

& 32
& 35.5
& 22.9
& 23.6
& 23.9
& 1.55$\times$
& 1.50$\times$
& 1.49$\times$ \\

\bottomrule
\end{tabular}

\end{table}

\section{Conclusion}
\label{sec:conclusion}
In this work, we presented \method{}, a low-overhead framework for MXFP4 quantization. Our approach addresses the weight quantization errors amplified by activation scaling, which concentrate in a small set of input channels of the weight matrix. \method{} combines AWQ-style scaling with a permutation that groups these channels into contiguous tail blocks and can be fused into the preceding projections offline. We precompute and quantize the corresponding weight residuals offline. At inference time, we copy the quantized tail activations together with their block scales, or obtain the augmented activations through offline producer expansion. This allows the main computation and residual compensation to share a single MXFP4 GEMM without online residual construction, irregular gathers, or specialized mixed-precision kernels. Experiments on both MoE and dense models show that applying \method{} only to FFN down-projections improves average downstream accuracy over default MXFP4.  Meanwhile, \method{} substantially accelerates the down-projection over BF16 and ARCQuant, while retaining end-to-end decoding efficiency close to that of plain MXFP4. These results demonstrate the benefits of structuring residual compensation around standard low-precision computation. Future work will explore extending \method{} beyond FFN down-projections to other Transformer layers, such as attention output and FFN up/gate projections, through specific offline fusion strategies.

\section*{Ethics Statement}
This work focuses on model compression and inference efficiency and does not involve new data collection or additional risks to user privacy. Low-bit quantization may alter model output behavior; therefore, task-level and safety regression tests should be conducted before deployment.

\section*{AI Use Statement}
We used AI solely to assist with language editing, including grammar correction and improvements to the clarity and readability of the manuscript. All AI-assisted edits were reviewed and revised by the authors to ensure that they accurately reflected the intended meaning. The authors take full responsibility for the final content of the paper.

\bibliographystyle{iclr2027_conference}
\bibliography{references}

\appendix

\section{Applicability and future work}

Although this work uses permutation as the primary example, the underlying principle is more general. Any channel-side operation acting on a hidden dimension shared by adjacent modules can support a low-overhead compensation mechanism, provided that it can be absorbed offline into the producer or consumer weights, or that it exposes the required activation sub-tensor through regular access. Because different layers provide different fusion opportunities, the applicable operations need not be restricted to permutation. They may also include diagonal scaling, block permutation, channel duplication, structured augmentation, or structured pruning with residual correction.

This principle applies most naturally to the second projection of an FFN, including the down-projection of a dense FFN and the down-projection of an MoE expert. For these layers, channel reordering, scaling, and tail duplication can be implemented by modifying adjacent weights offline. Similar ideas may also be extended to the attention output projection.

We select the MoE expert down-projection as the primary experimental setting for two reasons. First, this layer follows the gated activation, and prior work has observed token-local massive outliers at FFN down-projection inputs. SmoothQuant may also migrate activation-side difficulty at this position into pronounced weight-side outliers~\citep{lin2024duquant}. Second, existing MXFP4 methods typically rely on block-aware rotations, affine transformations, fallback paths, or residual/low-rank compensation to recover accuracy. These mechanisms may still introduce online activation transformations, dynamic residual construction, additional high-precision branches, or specialized kernel fusion~\citep{shao2025brq,lin2026duquantpp,li2026batquant,zhang2026osc,meng2026arcquant,li2024svdquant}. The MoE expert down-projection therefore combines substantial quantization difficulty with a suitable channel structure, making it a representative layer for evaluating the proposed method.

\end{document}